# Identifying depression-related topics in smartphone-collected free-response speech recordings using an automatic speech recognition system and a deep learning topic model

*Yuezhou Zhang[1,]*, Amos A Folarin[1,2,3,4], Judith Dineley[1,5], Pauline Conde[1], Valeria de Angel[1], Shaoxiong Sun[1], Yatharth Ranjan[1], Zulqarnain Rashid[1], Callum Stewart[1], Petroula Laiou[1], Heet Sankesara[1], Linglong Qian[1], Faith Matcham[1,6], Katie White[1], Carolin Oetzmann[1], Femke Lamers[7], Sara Siddi[8], Sara Simblett[1], Björn W. Schuller[5], Srinivasan Vairavan[8], Til Wykes[1,3], Josep Maria Haro[8], Brenda WJH Penninx[11], Vaibhav A Narayan[9,12], Matthew Hotopf[1,3], Richard JB Dobson[1,2,3,4,#], Nicholas Cummins[1,#,]*, RADAR-CNS consortium[13]*

[1]King's College London, London, UK
[2]University College London, London, UK
[3]South London and Maudsley NHS Foundation Trust, London, UK
[4]Health Data Research UK London, University College London, London, UK
[5]University of Augsburg, Augsburg, Germany
[6]School of Psychology, University of Sussex, Falmer, East Sussex, UK
[7]Department of Psychiatry and Amsterdam Public Health Research Institute, Amsterdam UMC, Vrije Universiteit, Amsterdam, The Netherlands
[8]Parc Sanitari Sant Joan de Déu, Fundació Sant Joan de Déu, CIBERSAM, Universitat de Barcelona, Barcelona, Spain
[9]Janssen Research and Development LLC, Titusville, NJ, USA
[11]Amsterdam University Medical Centre, Vrije Universiteit and GGZ inGeest, Amsterdam, Netherlands
[12]Davos Alzheimer's Collaborative, Geneva, Switzerland.
[13]www.radar-cns.org
[#]These authors contributed equally
*Corresponding authors

Emails of corresponding authors: yuezhou.zhang@kcl.ac.uk and nick.cummins@kcl.ac.uk

**Abstract**

Prior research has demonstrated correlations between language use and depression; however, large-scale validation is still needed. In conventional clinical practice, collecting spontaneous speech and written text samples is expensive and time-consuming. To address this challenge, many studies have applied natural language processing to social media posts to predict depression, but limitations remain, including a lack of validated depression labels, biased sampling of social media users, and no contextual information.

This study aimed to identify depression-related topics in smartphone-collected free-response speech recordings. Using an automatic speech recognition tool and topic modeling, we analyzed 3,919 speech recordings from 265 participants and identified 6 risk topics (out of 29) that were associated with higher depression severity: 'No Expectations', 'Sleep', 'Mental Therapy', 'Haircut', 'Studying', and 'Coursework'. Behavioral (from wearables) and linguistic characteristics were compared across all topics to understand the context of the identified topics. Overall, participants mentioning depression risk topics had higher sleep variability, later sleep onset, and fewer daily steps, and they spoke less, used more negative words, and mentioned leisure less during their speech recordings. Correlations between topic shifts and changes in depression severity over time were also investigated, indicating the importance of longitudinal language monitoring. Additionally, we validated these findings on a separate smaller dataset (356 speech recordings from 57 participants), obtaining some consistent results.

In summary, our findings demonstrate specific speech topics may indicate depression severity, though further validation is needed. The presented data-driven workflow provides a practical approach for analyzing large-scale speech data collected from real-world settings for digital health research.

# Introduction

Depression is one of the most prevalent mental health disorders[1], associated with adverse outcomes including premature mortality, diminished quality of life, disability, productivity loss, and suicide[2-5]. Its chronic nature, poor prognosis, and comorbidities impose substantial economic and societal burdens[6,7]. The current clinical diagnosis of depression relies heavily on subjective recall and the expertise of skilled clinicians using questionnaires and interviews[8,9], leading to inadequate and delayed treatment for depressed individuals[10]. More effective and objective methodologies are therefore needed to identify depression at an early stage[11].

Spoken language can reflect an individual's personality, emotions, and mental health status[12]. The relationship between changes in language and depression has been demonstrated previously, for example, the increased use of negative words and first-person pronouns was found to be associated with worsening depression symptoms[13,14]. These relationships need to be validated in large-scale studies. However, in conventional clinical practice, collecting language samples (e.g., spontaneous speech and written essays) is time-consuming and expensive, resulting in small sample sizes[12]. To address this limitation, recent studies have applied natural language processing (NLP) technologies to a huge volume of posts from social media, such as Twitter, Facebook, and Reddit, to predict depression[15-17]. Specific discussion topics in posts, such as low mood, loneliness, hostility, somatic complaints, and medical references, have been found to be significantly correlated with depression[18].

These social media-based studies also have limitations. Firstly, the samples collected cannot be regarded as clinical, most used depression-related keywords or tags as the depression label rather than a clinically valid diagnosis/questionnaire of depression. Second, language data is restricted to social media users only, lacking language information from non-social media users, which biases the findings. Third, contextual information (such as sleep and activity) was not collected, limiting the understanding of the potential reasons why certain topics emerge.

The development of mobile and Internet-of-Things technologies provides a cost-efficient means to concurrently track individuals' spoken language via speech recordings, depression status via remote questionnaires, and behavioral information via wearable devices[19-21]. Therefore, these technologies provide us opportunities to link spoken language use with depression in real-world settings, within the context of behaviors[11,22]. For instance, if a participant expressed looking forward to better sleep in a speech recording, wearable-measured sleep metrics can be used to investigate the sleep quality of the participant before the speech task. Also, behavioral characteristics were reported to be significantly associated with depression severity[23-27], which can aid in understanding the depressive state of participants when conducting speech recordings.

A major bottleneck in analyzing speech recordings collected in mobile health studies is the manual transcription of speech recordings, the time and cost burdens of which are not realistic for use in real-world mHealth pipelines. Recent developments in automatic speech recognition (ASR) systems, such as OpenAI's Whisper[28], achieve near human-level robustness and accuracy, enabling large-scale transcription of speech-to-text and then NLP to extract language patterns.

The primary aims of this study were to (i) utilize a data-driven workflow to identify language topics associated with depression severity from free-response speech recordings collected from a clinical population via smartphones in daily-life settings; (ii) explore differences in contextual behaviors (extracted from wearables) and linguistic characteristics across identified topics to understand participants' mental health status before/during the speech tasks and speculate reasons for topic emergence; (iii) explore whether changes in speech topics correlate to changes in depression severity. We performed a secondary analysis on the Remote Assessment of Disease and Relapse Major Depressive Disorder (RADAR-MDD) dataset[29,30], an observational and longitudinal depression study collecting multimodal data streams, including the free-response speech task (allowing participants to describe what they were looking forward to in the next week), depression questionnaire (the 8-item Patient Health Questionnaire [PHQ-8][31]), and wearable data. A deep learning-based topic model, BERTopic[32], was used to identify topics in the texts transcribed from speech recordings using the Whisper ASR tool. PHQ-8 scores and behavioral/linguistic characteristics were then compared across topics (Figure 1). Additionally, we investigated the topic shifts between two consecutive speech task and their correlation with changes in PHQ-8 scores. Furthermore, we validated the established topic model on a separate smaller mobile health dataset, Remote Assessment of Treatment Prognosis in Depression (RAPID)[33,34], to test the generalizability of our findings.

**Figure 1.** The data analytical workflow. It contains 4 steps: (1) free-response speech recording collection, (2) speech-to-text transcription via Whisper, (3) topic modeling using BERTopic, (4) comparisons of PHQ8 scores and behavioral/linguistic characteristics across identified topics to identify risk topics for depression.

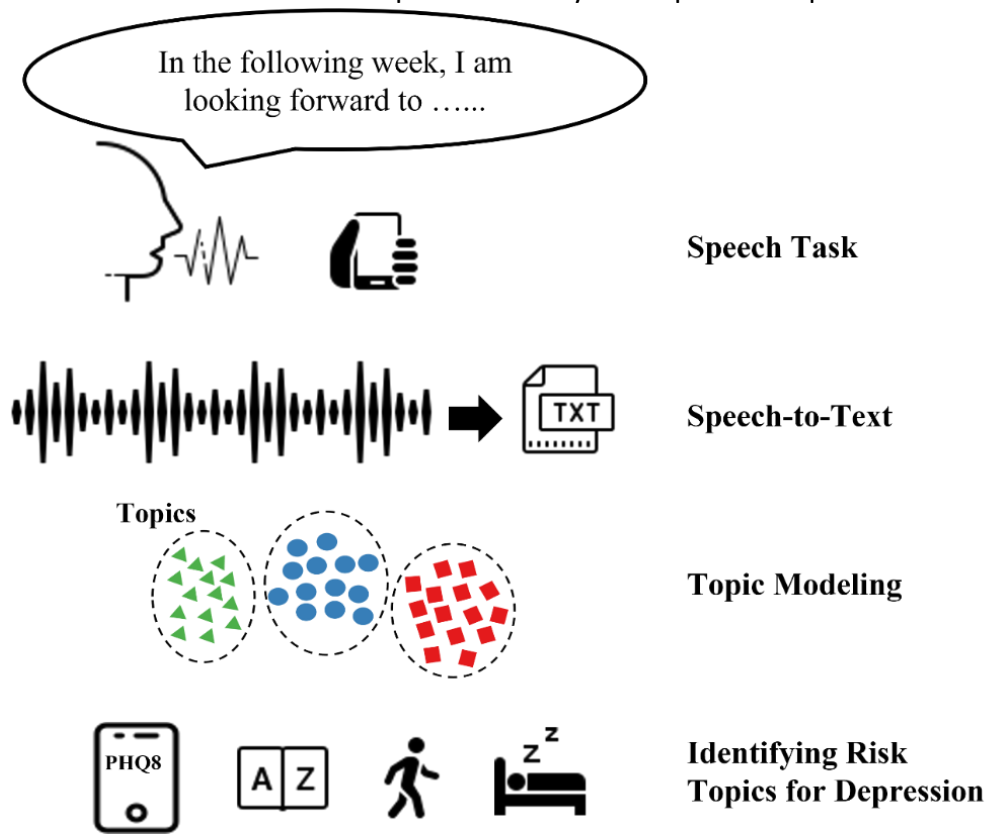

# Results

The free-response English speech recordings in RADAR-MDD were automatically transcribed into texts via OpenAI's Whisper tool[28]. After manually removing abnormal texts (see Methods), a total of 3,919 transcribed speech texts from 265 participants were analyzed in this study. The cohort had a median (IQR) age of 46.00 (31.00, 58.00) years and was predominantly female (78.5%) and White (88.3%). The detailed socio-demographics are provided in Supplementary Table 1.

**Topic Modeling Results**

We identified 29 topics in the transcribed texts using the BERTopic model (Table 1). We summarize these topics in 7 categories below.

Nothing to Expect: This category encompasses two topics, namely 'No Expectations' and 'No Expectations due to Covid'. Both topics revealed participants had nothing to look forward to in the coming week. Notably, the 'No Expectations' topic accounted for a considerable proportion, encompassing 302 recordings from 101 participants. The latter topic, 'No Expectations due to Covid', showed that the lack of expectations was attributed to COVID-19 restrictions, as exemplified by "Because of the lockdown and social isolation, I have nothing to look forward to over the next seven days."

Social Networks and Activities: Some participants' expectations for the next week were related to social networks and activities, including topics of 'Family' (keywords: daughter, son, parents), 'Friend' (keywords: friends, meeting, seeing), 'Festival' (keywords: Christmas, new year, Easter), 'Celebration' (keywords: birthday, celebrating, party), and 'Conversation' (keywords: phone, skype, chat).

Entertainment and Hobbies: The forthcoming entertainment plans of participants were manifest in topics of 'Art Activity' (keywords: rehearsal, pantomime, theatre), 'Holiday' (keywords: holiday, half term, away), 'Traveling' (keywords: scenery, train, trip), and 'Weekend' (keywords: weekend, day off, relaxing). Additionally, the topics of 'Gardening' (keywords: gardening, planting, plants) and 'Hobby' (keywords: yoga, baking, pottery) reflected participants' desires to engage in hobbies during the next week.

Study and Work: Participants' expectations regarding academic and work-related activities for the upcoming week were reflected in topics of 'Coursework' (keywords: university, course, exams), 'Studying' (keywords: book, reading, writing), 'Online Meeting' (keywords: zoom, quiz, meeting), and 'Working' (keywords: job, work, project).

Sports: Four distinct topics were associated with participants' sport-related plans for the coming week, including 'Fitness' (keywords: gym, exercise, fitness), 'Walking' (keywords: walk, walking, going), 'Outdoor Activity' (keywords: ride, climbing, cycling), and 'Swimming' (keywords: swimming, pool, sea).

Health: The topics of 'Hospital' (keywords: hospital, operation, pain), 'Mental Therapy' (keywords: mental, therapy, NHS), and 'Sleep' (keywords: sleep, tired, rest) reflect the health conditions of participants.

Other Themes: Additional topics included subjects such as 'Pet' (keywords: dog, puppy, cat), 'Weather' (keywords: sunshine, rain, warm), 'House' (keywords: room, decorating, bedroom), 'Covid-19' (keywords: Covid-19, virus, vaccine), and 'Haircut' (keywords: haircut, hair, cut).

**Comparison between Topics**
**Depression Symptom Severity**
We observed significant variability in PHQ-8 scores across the topics (Kruskal-Wallis test: $P<.001$) (Figure 2 and Supplementary Table 2). There were six topics with a median PHQ-8 score $\geq$ 10 (depression screening threshold), including 'No

Expectations' (13.00 [8.00, 18.00]), 'Sleep' (13.00 [7.75, 15.25]), 'Mental Therapy' (12.50 [5.00, 15.75]), 'Haircut' (11.00 [7.50, 15.50]), 'Studying' (11.00 [7.50, 14.50]), and 'Coursework' (10.50 [6.75, 14.00]). These six topics were regarded as risk topics for depression. Other topics with median PHQ-8 scores ranging from 5 to 9 were regarded as non-risk topics for depression such as 'Art Activity' (6.00 [3.00, 12.00]), 'Gardening' (6.00 [3.00, 11.00]), 'Holiday' (5.00 [2.00, 9.00]), and 'Outdoor Activity' (5.00 [2.00, 8.00]).

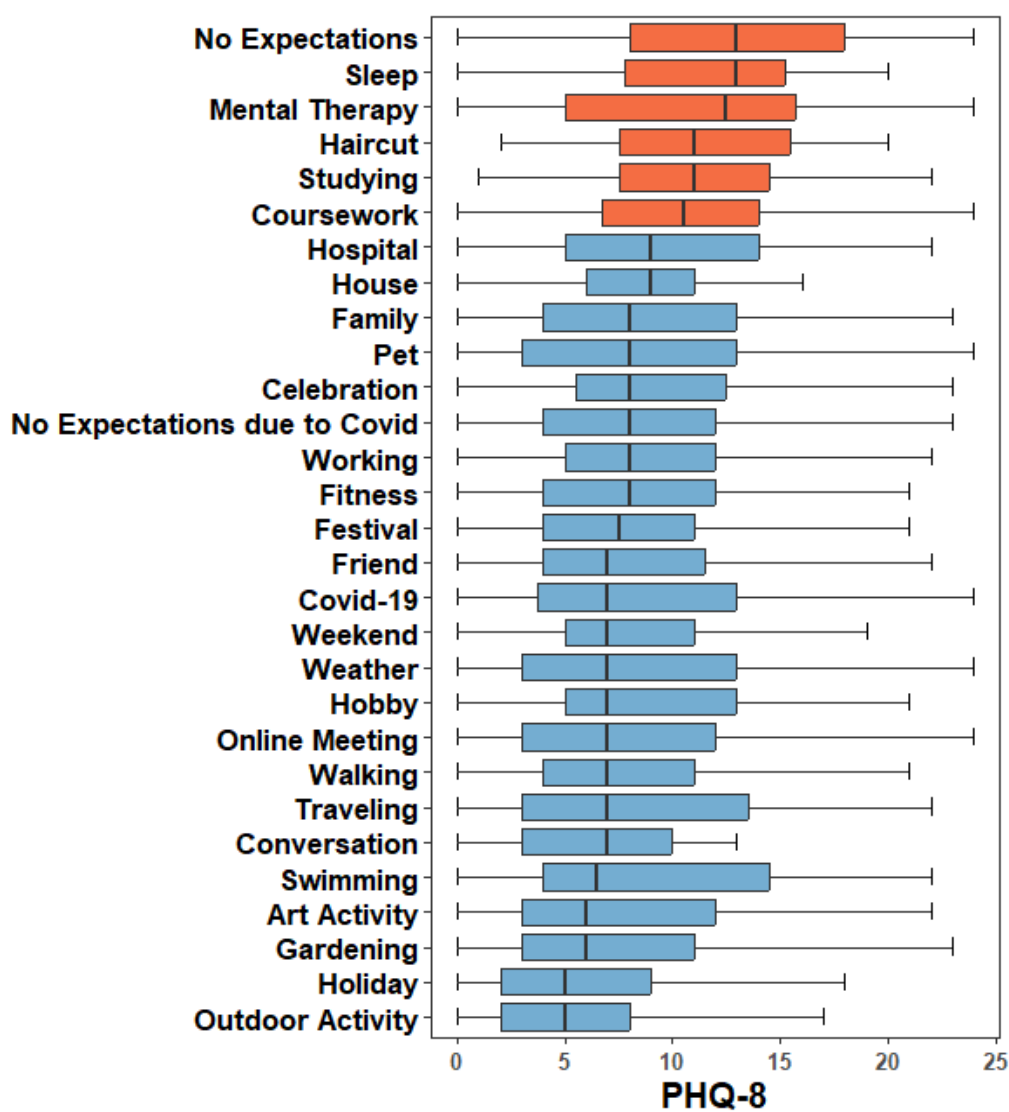

**Figure 2.** Boxplots of PHQ-8 scores for 29 topics identified in RADAR-MDD's free-response speech recordings. Note, orange represents risk topics (median PHQ-8 ≥ 10) and blue represents non-risk topics (median PHQ-8 < 10).

**Behavioral Characteristics**

We found that behavioral features derived from one-week Fitbit recordings prior to the speech task (see Method section) – *Sleep Variability*, *Sleep Onset*, and *Daily Steps* – were significantly different across topics (Kruskal-Wallis test: $P < .001$) (Figure 3 and Supplementary Table 3).

Participants who mentioned risk topics, except for the 'Studying' topic, had higher sleep variability over the preceding week compared to those discussing non-risk topics. Participants discussing the 'Sleep' topic had the highest sleep variability (1.65 [1.33, 1.78] hours), while participants who mentioned the 'Outdoor Activity' topic had the lowest sleep variability (0.67 [0.37, 0.80] hours) (Figure 3a). For sleep onset time, participants who mentioned 'Sleep' (01:16 [23:37, 03:19]), 'Coursework' (00:40 [23:18, 01:54]), and 'No Expectations' (00:28 [23:22, 02:03]) topics slept later than participants mentioning the non-risk topics (23:53 [22:58, 00:59]) (Figure 3b).

Regarding daily activities, participants mentioning risk topics for depression (except for the 'Haircut' topic) had fewer daily steps than participants who talked about non-risk topics. Specifically, participants discussing 'Coursework' (3202.57 [1741.29, 6569.86] steps), 'No Expectations' (3923.00 [1156.50, 6589.07] steps), and 'Sleep' (3899.71 [3036.14, 6951.71] steps) had relatively low daily step count, whereas participants mentioning the 'Holiday' topic had the most daily steps (7747.14 [5269.71, 11147.29] steps) (Figure 3c).

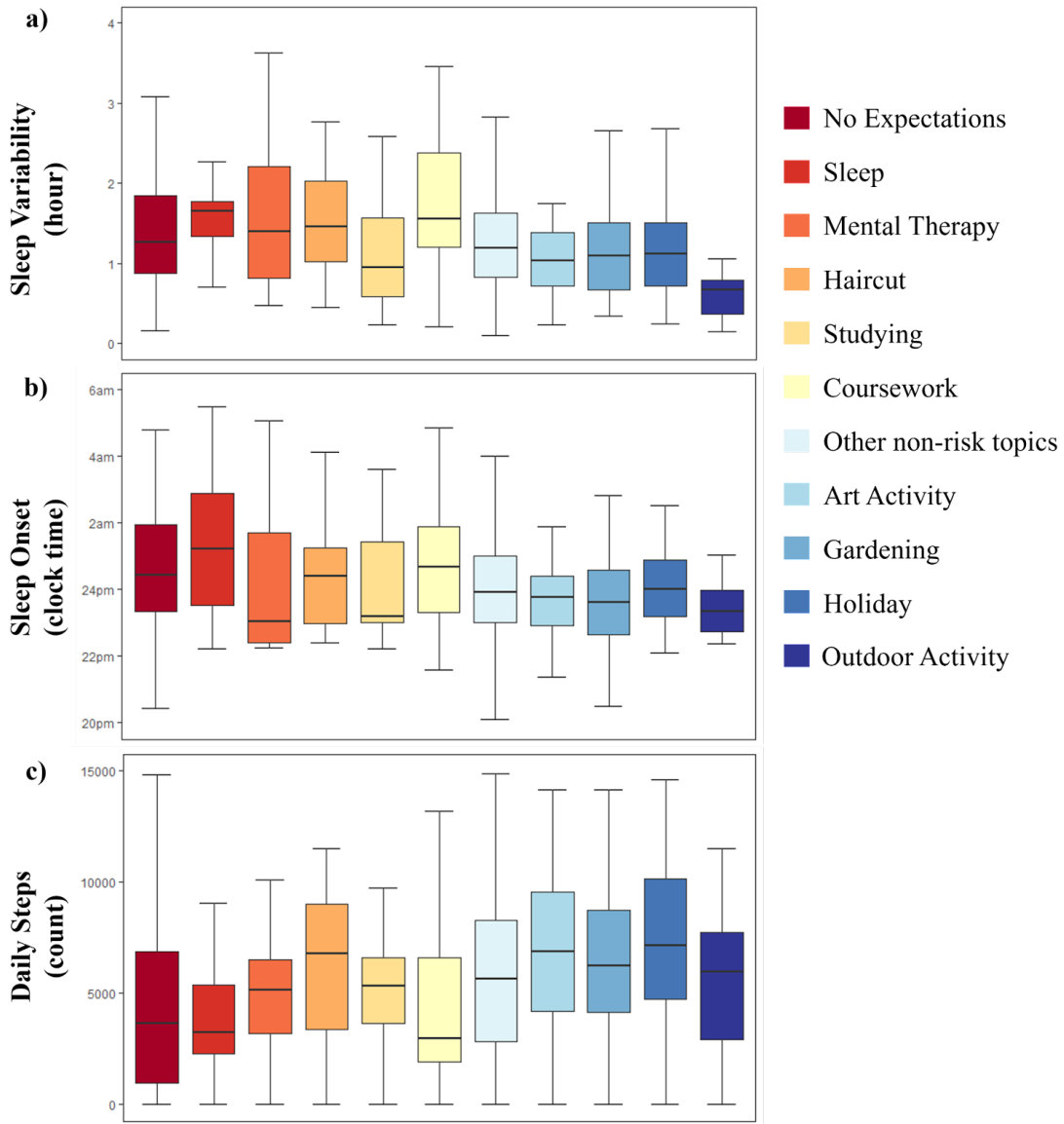

**Figure 3.** Boxplots of behavioral features for 6 identified risk topics, 4 non-risk topics with the lowest median PHQ-8 scores ('Art Activity', 'Gardening', 'Holiday', and 'Outdoor Activity'), and other non-risk topics (grouped as a single category).

**Linguistic Characteristics**

We also found several linguistic features (extracted by the Linguistic Inquiry and Word Count tool (LIWC-22)[35]; see Method and Supplementary Table 4) were significantly different across topics (Kruskal-Wallis test: *P* < .001). We observed four notable differences in *Word Count*, *Negation*, *Leisure*, and *Negative Emotion* between risk and non-risk topics for depression (Figure 4). Further details of comparisons are provided in Supplementary Table 5.

In the speech tasks, participants discussing 'Mental Therapy' (40.00 [28.50, 66.50] words) and 'Holiday' (36.00 [16.00, 71.00] words) topics spoke a greater number of words, while participants mentioning 'Studying' (13.00 [10.50, 21.50] words) and 'No Expectations' (13.00 [10.00, 23.00] words) topics spoke less (Figure 4a). Participants expressing the 'No Expectations' topic used the highest percentage of negation terms in their speech tasks (90.1%) compared to other participants (Figure 4b). Also, participants used more leisure-related words when they were talking about topics of 'Art Activity' (68.0%), 'Gardening' (77.9%), and 'Holiday' (55.7%), whereas participants who mentioned 'No Expectations' (2.6%) and 'Haircut' (4.3%) topics used fewer leisure-related words (Figure 4c). Furthermore, speech contents of the 'Sleep' and 'Mental Therapy' topics contained more negative emotion words, while speech texts of the 'Studying' and 'Art Activity topics' contained fewer negative emotion words (Figure 4d).

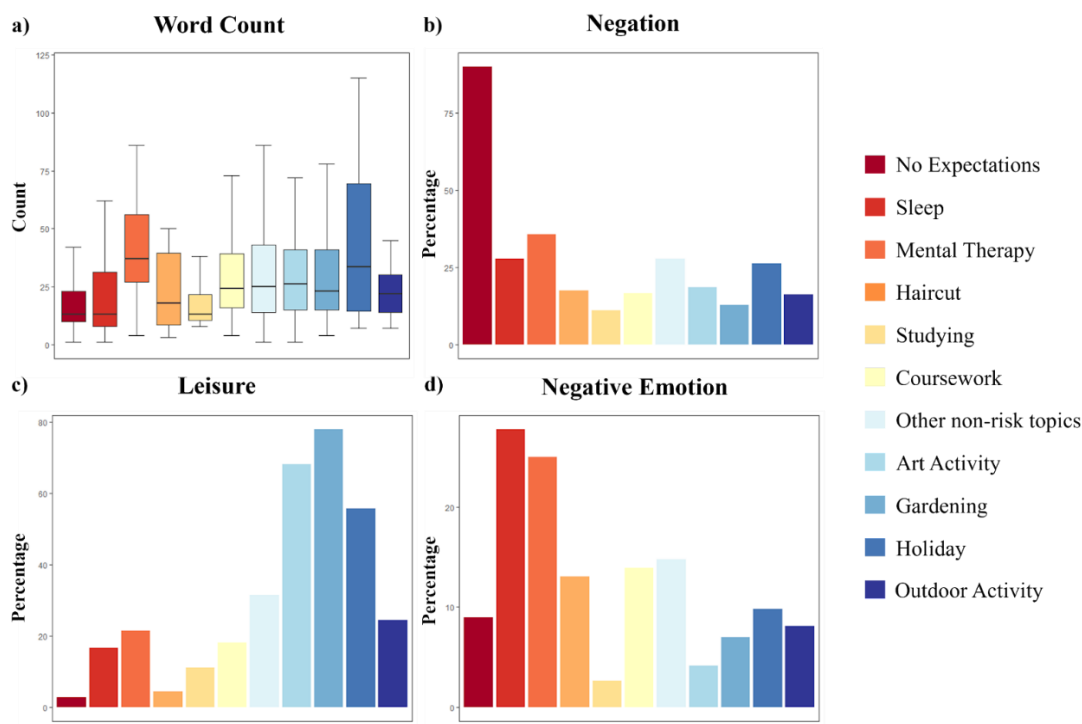

**Figure 4.** Comparison of linguistic features of Word Count, Negation, Leisure, and Negative Emotion across 6 risk topics, 4 non-risk topics with the lowest median PHQ-8 scores ('Art Activity', 'Gardening', 'Holiday', and 'Outdoor Activity'), and other non-risk topics (grouped as one category).

**Topic Shifts and Changes in Depression Severity Over Time**

There are four topic shifts between two consecutive speech tasks: Risk to Risk, Risk to Non-Risk, Non-Risk to Risk, and Non-Risk to Non-Risk. We observed that the PHQ-8 difference of the Risk to Non-Risk shift was significant (Kruskal-Wallis test: $P<0.001$), indicating if the speech topic changed from the risk topic to the non-risk topic, the corresponding PHQ-8 score decreased by 1.1 points on average (Figure 5a). We also observed that the PHQ-8 scores of two consecutive risk topics (Risk to Risk) were higher than those with only one risk topic in two consecutive speech tasks (Risk to Non-Risk and Non-Risk to Risk) (Figure 5b). After/before a risk topic, the PHQ-8 was still higher even if the corresponding speech topic was non-risk (Risk to Non-Risk: 10.0 [6.0, 15.0] and Non-Risk to Risk: 11.0[6.0,16.0]), compared to those of two consecutive non-risk topics (7.0 [4.0, 12.0]) (Kruskal-Wallis test: $P<0.001$) (Figure 5b).

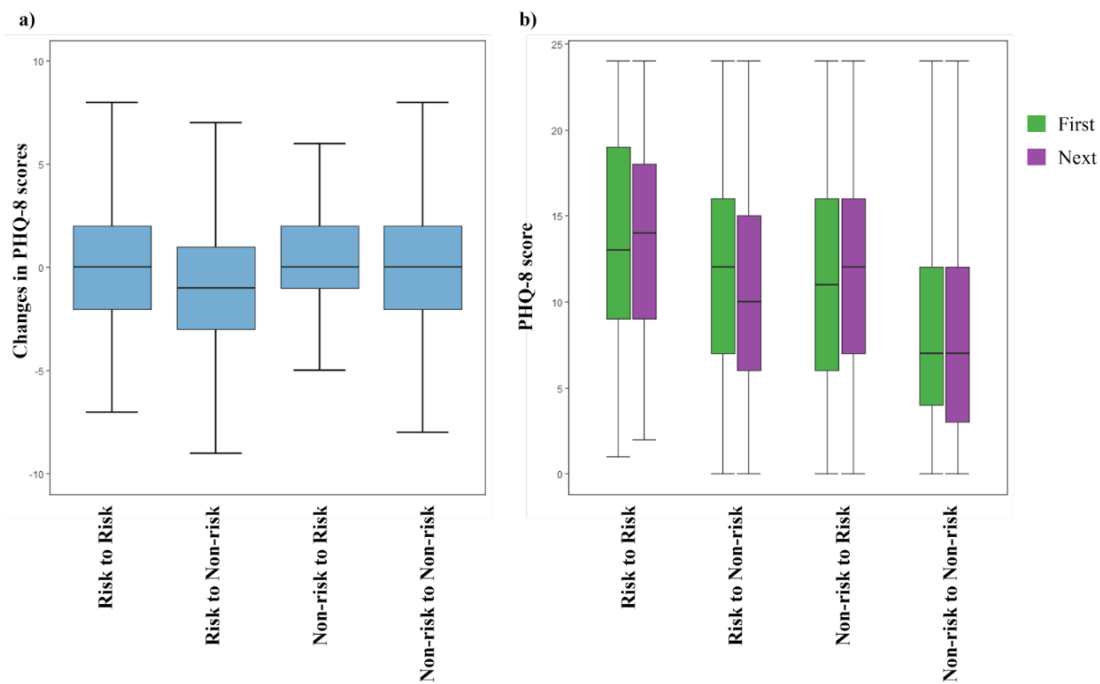

**Figure 5.** The correlation between topic shifts and changes in depression severity. a) The boxplots of changes in PHQ-8 scores between two consecutive speech tasks for the 4 topic shifts. b) boxplots of PHQ-8 scores corresponding to two consecutive speech tasks for the 4 topic shifts. Note, there are 4 topic shifts between two consecutive speech tasks: Risk to Risk, Risk to Non-Risk, Non-Risk to Risk, and Non-Risk to Non-Risk.

**Validation on the RAPID Dataset**

After manually removing abnormal speech tasks, a total of 356 transcribed speech texts from 57 participants were available in the RAPID dataset. We used the established BERTopic model to classify these speech texts into topics (Supplementary Table 6). Due to the small sample size of the dataset, some topics contained a limited number of speech texts. Therefore, we only compared the 9-item Patient Health Questionnaire (PHQ-9[36]) scores across topics comprising a minimum of 10 transcribed speech texts (Figure 6). We found that participants who mentioned 'No Expectations' and 'Sleep' topics had significantly higher median PHQ-9 scores (17.00 [15.00, 19.00] and 18.00 [15.00, 19.00]) compared to those discussing other topics (Kruskal-Wallis test: *P*=.006).

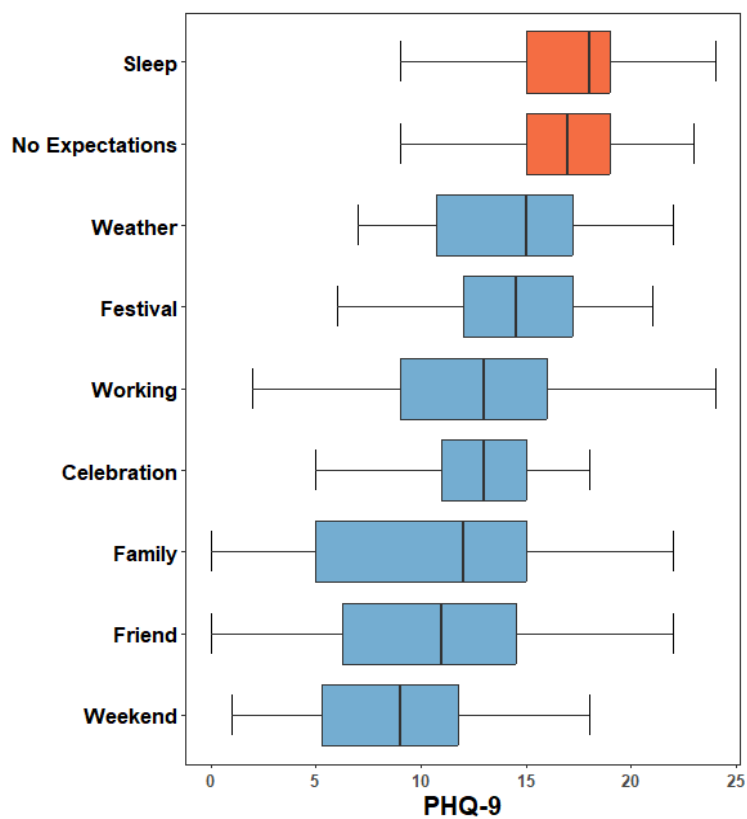

**Figure 6.** Boxplots of PHQ-9 scores for topics in RAPID speech tasks using the established BERTopic model (built on RADAR-MDD dataset). Note, topics with at least 10 speech texts are displayed. Orange represents risk topics and blue represents non-risk topics for depression that were identified in the RADAR-MDD dataset.

# Discussion

Utilizing automatic speech recognition, topic modeling, and mobile technologies, the present study proposed a data-driven approach for analyzing large-scale spoken language data. To the best of our knowledge, it is the first study to explore depression-related language topics on real-world speech recordings collected from two different clinical populations in their daily life. Here, we discuss our key findings alongside prior research.

**Risk Topics for Depression**

A notable finding is a considerable proportion of participants with high depression severity expressed having nothing to look forward to in the upcoming week in their free-response speech tasks, which is consistent in both datasets. This could be attributed to hopelessness being a key facet of depression[37,38]. The incapacity to hope is frequently found in first-person reports of depressed people[39]. Prior literature described such sentiments of depression as "all sense of hope had vanished", "a paralysis of hope", and "no words to explain the depths of my despair"[40,41]. Moreover, previous quantitative studies reported a significant correlation between hopelessness (measured by questionnaires) and depression[42,43]. Thus, the expression of hopelessness for the future may serve as an indicative marker of depression.

Participants expressing the expectation of sleep had relatively high depression severity, this finding is also consistent in both datasets. The potential reason is that participants looking forward to sleep may have suffered from poor sleep or been exhausted before their speech tasks. It is known that sleep disturbances are key manifestations of depression[44,45]. Further, fatigue is also a somatic symptom of depression[46]. Therefore, mentioning the expectation of sleep may be also an indicator of depression.

Many of our participants who mentioned studying, reading, examinations, and coursework had relatively high depression symptom severity. Several studies reported that workload[47-49], exams[50,51], and concerns about academic performance[52,53] are risk factors for stress, anxiety, and depression among college students. Since the 'Mental Therapy' topic is directly related to mental disorders, participants who mentioned this topic were likely experiencing mental health problems. Intriguingly, 21 participants who mentioned the 'Haircut' topic also had relatively high depression severity. However, due to a lack of contextual information on exactly why topics related to haircuts were being raised and limited sample size, this specific link needs future investigation and validation on large datasets.

**Non-risk Topics for Depression**

Participants discussing topics related to art activities, gardening, holidays, and outdoor activities had the lowest depression severity in our cohort. Engaging in art activities, such as dance, singing, and theatrical performances, has been reported to have beneficial effects on individuals' mental health[54-56]. Moreover, since exposure to nature can promote feelings of connectedness and relaxation[57], participating in gardening activities has been linked to reduced depression and anxiety symptoms, as well as increased positive emotions[58,59]. Similarly, holiday trips have been found to contribute to lessening loneliness[60], enhancing life satisfaction[61], and alleviating depression[62]. Furthermore, engaging in outdoor activities like walking, running, and cycling, has demonstrated a notable association with a reduced risk of depression[63].

For some other non-risk topics detected in the RADAR-MDD, including 'Conversation', 'Walking', and 'Fitness', previous studies have reported that social and physical activities are negatively associated with depression[64,65].

**Topics and Passive Behavioral Characteristics**

To the best of our knowledge, this study is the first to link spoken language topics with matching behavioral information as measured by wearable devices. Passive behavioral features extracted from Fitbit recordings could help understand the potential emergence reason for some topics and the depressive status of participants when conducting speech tasks. Using this information, we were able to observe that participants expressing the 'Sleep' topic had the highest sleep variability and the latest sleep onset time during the week before their speech tasks, indicating that they were more likely to be experiencing unstable sleep or insomnia[23]. Also, participants who mentioned 'Coursework' had late and disrupted sleep during the past week, likely due to academic workload and study pressure. In contrast, participants who expressed non-risk topics slept relatively better, indicating that their mental health was likely in better status. Regarding daily activities, participants expressing 'No Expectations', 'Sleep', and 'Coursework' topics had fewer daily step counts than those mentioning other topics which may be caused by a lack of motivation, fatigue, and pressure of academic workload, respectively. The low activity levels of these participants indicated that they may have had relatively high severity of depression symptoms before speech tasks[24].

**Topics and Linguistic Characteristics**

The linguistic characteristics of the transcribed speech texts could provide additional insights into the participants' mental status (e.g., emotion) during the speech task. We observed several significant differences in linguistic features across topics, which provide support to our findings. In one of our previous studies, we found participants with higher depression severity tended to speak less in their speech tasks[66]. This may explain why the speech texts of some risk topics ('No Expectations', 'Sleep', and 'Studying') contain fewer words than those of non-risk topics. Furthermore, prior research shows that engagement in leisure activities can be a protective factor against depression[67,68]. This aligns with our observation that participants discussing non-risk topics used more leisure-related words than those mentioning risk topics. Additionally, previous studies have linked elevated use of negative emotion words with higher depression severity[69,70]. In our data, participants who mentioned 'Sleep' and 'Mental Therapy' topics used more negative emotional words than other topics, potentially indicating their depressed states. Moreover, participants who mentioned the 'No Expectations' topic were more likely to use negated words (such as nothing, don't, and can't) when expressing that they had nothing to look forward to in the next week.

**Topic Shifts and Changes in Depression Severity Over Time**

To the best of our knowledge, there is currently a lack of longitudinal research tracking how the changes in spoken language topics relate to fluctuations in depression severity over time. We found that participants had higher depression severity when mentioning two consecutive risk topics than those who mentioned only one risk topic within two consecutive speech tasks. We also found that before or after a risk topic, depression severity was still higher even if the corresponding topic was a non-risk topic, compared to when two consecutive topics were both non-risk topics. Furthermore, changing speech topics from risk to non-risk was significantly associated with a decrease in PHQ-8 scores. These findings indicate that risk topics may reflect the depressive status of a long period and frequently discussing them may reflect severe depression severity, which highlights the importance of longitudinally monitoring language.

**Limitations and Future Research**

Our findings should be interpreted in the context of certain limitations related to our cohorts and the format of the free-response speech task. First, participants in the present study had a history of major depression (RADAR-MDD) or were pursuing a depression treatment (RAPID). Our results may not generalize to non-clinical populations. Additionally, some topics' sample sizes were small and required validation in larger corpora. Our findings were based on a specific speech task, describing one's expectations for the upcoming week. Thus, our findings may not be generalizable to other free-response speech tasks.

Participants may have different motives when mentioning a certain topic. For instance, some participants mentioned no expectations because they had nothing special things to expect or preferred not elaborating, rather than due to hopelessness per se. Thus, this study only explored general associations between speech topics and depression. Further analysis incorporating additional information is needed to distinguish different situations and motives.

The transcription of spoken language was performed with a pre-trained deep learning model that may contain errors; depression affects the acoustic properties of speech[71] which may alter the accuracy of the transcripts. Further investigations are needed for these issues. Nevertheless, we were still able to identify several risk topics for depression that match with related findings in the literature and the use of ASR provides a practical way for analyzing a huge volume of real-world speech data from large-scale mobile health studies.

## Method

### Datasets

**The RADAR-MDD Dataset**

The RADAR-MDD study is a large observational investigation assessing the utility of remote technologies for monitoring depression[29]. The study recruited 623 participants with a depression history from three sites in the United Kingdom, Spain, and the Netherlands, and followed them for up to over 2 years[29]. Enrollment began in November 2017, ended in June 2020, and data collection finished in April 2021[30]. The study used RADAR-base[72], an open-source platform, to remotely gather participants' active (questionnaires and speech tasks) and passive (smartphones and Fitbit) data streams.

Speech recordings involved two tasks: (i) a scripted task, reading a script from Aesop's fable The North Wind and The Sun[73], and (ii) a free-response task, allowing participants to describe what they were looking forward to in the upcoming week[71,74]. Details of the RADAR-MDD dataset and speech data collection have been reported in our previous publications[21,29,30,66]. This study focused only on English free-response speech recordings from the UK site due to our research aims and the performance variation of Whisper in language translation.

**Patient Involvement**

The RADAR-MDD protocol was co-developed with a patient advisory board who shared their opinions on several user-facing aspects of the study including the choice and frequency of survey measures, the usability of the study app, participant-facing documents, selection of optimal participation incentives, selection, and deployment of wearable device as well as the data analysis plan. The speech tasks and subsequent analysis have been discussed specifically with the RADAR-CNS *Patient Advisory Board* (PAB).

**The RAPID Dataset**

The RAPID study investigated the feasibility of remote technologies during the treatment of depression[33,34]. From June 2020 to June 2021, 66 adults in London, UK seeking depression treatments were recruited and followed for 7 months[33,34]. Since the RAPID used the same speech collection protocol as the RADAR-MDD, the free-response speech recordings collected in the RAPID were used for validation.

**Ethics Considerations**

The RADAR-MDD and RAPID studies were both conducted per the Declaration of Helsinki and Good Clinical Practice, adhering to principles outlined in the National Health Service (NHS) Research Governance Framework for Health and Social Care (2nd edition). The ethical approval of the RADAR-MDD of the UK site had been obtained in London from the Camberwell St Giles Research Ethics Committee (REC reference: 17/LO/1154). The RAPID study was reviewed by the London Westminster Research Ethics Committee and received approval from the Health Research Authority (reference number 20/LO/0091). All participants of both studies signed informed consent.

**Data Processing and Analysis**

We first utilized Whisper[28] to automatically the speech recordings into texts. Then, we applied the BERTopic model[32] to identify primary topics in transcribed texts. To identify depression risk topics and understand the topic context, we compared the depression symptom severity and behavioral/linguistic characteristics across identified topics (Figure 1).

**Speech-to-Text Transcription**

Whisper, an OpenAI automatic speech recognition system (ASR) trained on 680,000 hours of web-collected audio data, outperformed the state-of-the-art speech recognition systems on public speech datasets covering various domains, tasks, and languages[28]. We have previously used Whisper to transcribe smartphone-collected speech samples[75]. In this study, we utilized the Whisper Medium model (https://github.com/openai/whisper) to transcribe our free-response speech recordings into texts After transcription, we manually reviewed the texts and removed abnormalities that were caused by technical or operational issues (e.g., empty recordings).

**Topic Modeling**

BERTopic[32], a deep learning-based topic model, was chosen to identify principal topics in the transcribed texts due to its contextual understanding and efficiency with short texts[76,77]. Then, we assigned an appropriate title to represent each of the identified topics by reviewing keywords generated by BERTopic.

**Comparisons between Topics**

We summarized and compared depression symptom severity and behavioral/linguistic features across identified topics. Differences between topics were assessed for significance using Kruskal-Wallis tests[78]. These are briefly described below.

Depression Symptom Severity. The participant's depression symptom severity was measured biweekly using the 8-item Patient Health Questionnaire (PHQ-8)[31] (PHQ-9[36] for RAPID) along with the speech task. A PHQ-8 $\geq$ 10 is the recommended threshold for depression screening[31]. Therefore, the topics mentioned by participants with a median PHQ-8 $\geq$ 10 were considered as the risk topics for depression; other topics were regarded as non-risk topics.

Behavioral features. To understand the potential reasons for topic emergence, from 1-week Fitbit data preceding the speech task, we extracted the following three behavioral features: *Sleep Variability* (standard deviation of sleep duration over a period), *Sleep Onset* (mean sleep onset time over a period), and *Daily Steps* (mean steps per day), as they have been reported to be significantly associated with depression severity[23,24].

Linguistic features. Based on prior literature linking depression and language[69,70,79,80], we extracted 20 linguistic features, including some summary variables (e.g., word count), personal pronouns, emotional words, and lifestyle, from the transcribed texts using the latest version of the Linguistic Inquiry and Word Count tool (LIWC-22)[35]. The extracted linguistic features are listed in Supplementary Table 4.

**Topic Shifts and Changes in Depression Severity Over Time**

This study also aimed to examine potential associations between speech topic shifts and changes in depression severity. There are four types of topic shifts between two consecutive speech tasks: Risk to Risk, Risk to Non-Risk, Non-Risk to Risk, and Non-Risk to Non-Risk. We utilized the Kruskal-Wallis test[78] to assess whether differences in PHQ-8 scores between the two speech tasks are significantly different across these four topic shift patterns.

## Data availability

The datasets used for the present study can be made available through reasonable requests to the RADAR-CNS consortium. Please email the corresponding author for details.

## Code availability

The complete code used for the analysis can be made available through reasonable requests to the RADAR-CNS consortium. Please email the corresponding author for details.

## Author's contributions

Y.Z., N.C., A.F., and R.D. contributed to the design of the study. Y.Z. performed the analysis and drafted the manuscript. M.H. is the principal investigator for the RADAR-MDD study. N.C., J.D., P.C., F.M., K.W., C.O., F.L., S. Siddi, S. Simblett, J.M.H., B.W.J.H.P., and M.H. contributed to participant recruitment and data collection. R.D., A.F., Y.R., Z.R., P.C., H.S., and C.S. have contributed to the development of the RADAR-base platform for data collection. A.F., V.A.N., T.W., M.H., and R.D. contributed the administrative, technical, and clinical support of the study. All authors were involved in reviewing the manuscript, had access to the study data, and provided direction and comments on the manuscript.

## Competing interests

S.V. and V.A.N. are employees of Janssen Research and Development LLC. M.H. is the principal investigator of the Remote Assessment of Disease and Relapse–Central Nervous System project, a private public precompetitive consortium that receives funding from Janssen, UCB, Lundbeck, MSD, and Biogen.

**Acknowledgments**

The RADAR-CNS project has received funding from the Innovative Medicines Initiative 2 Joint Undertaking under grant agreement No 115902. This Joint Undertaking receives support from the European Union's Horizon 2020 research and innovation programme and EFPIA, www.imi.europa.eu. This communication reflects the views of the RADAR-CNS consortium and neither IMI nor the European Union and EFPIA are liable for any use that may be made of the information contained herein. The funding body has not been involved in the design of the study, the collection or analysis of data, or the interpretation of data.

This paper represents independent research part-funded by the National Institute for Health Research (NIHR) Maudsley Biomedical Research Centre at South London and Maudsley NHS Foundation Trust and King's College London. The views expressed are those of the author(s) and not necessarily those of the NHS, the NIHR or the Department of Health and Social Care.

We thank all the members of the RADAR-CNS patient advisory board for their contribution to the device selection procedures, and their invaluable advice throughout the study protocol design.

This research was reviewed by a team with experience of mental health problems and their careers who have been specially trained to advise on research proposals and documentation through the Feasibility and Acceptability Support Team for Researchers (FAST-R): a free, confidential service in England provided by the National Institute for Health Research Maudsley Biomedical Research Centre via King's College London and South London and Maudsley NHS Foundation Trust.

**Table 1.** A summary of 29 topics identified in the free-response speech tasks of the RADAR-MDD dataset.

| Topic | Number of Recordings | Number of Participants | Keywords |
|---|---|---|---|
| **Nothing to Expect** | | | |
| No Expectations | 302 | 101 | nothing to look forward to |
| No Expectations due to Covid | 65 | 39 | lockdown, restrictions, nothing to look forward to |
| **Social Networks and Activities** | | | |
| Family | 289 | 123 | daughter, son, parents |
| Friend | 171 | 94 | friends, meeting, seeing |
| Festival | 154 | 101 | Christmas, new year, Easter |
| Celebration | 123 | 82 | birthday, celebrating, party |
| Conversation | 26 | 23 | phone, skype, chat |
| **Entertainment and Hobbies** | | | |
| Art Activity | 97 | 56 | rehearsal, pantomime, theatre |
| Holiday | 61 | 44 | holiday, half term, away |
| Traveling | 39 | 35 | scenery, train, trip |
| Weekend | 116 | 70 | weekend, day off, relaxing |
| Gardening | 86 | 55 | gardening, planting, plants |
| Hobby | 81 | 45 | yoga, baking, pottery |
| **Study and Work** | | | |
| Coursework | 72 | 38 | university, course, exams |
| Studying | 27 | 16 | book, reading, writing |
| Online Meeting | 63 | 36 | zoom, quiz, meeting |
| Working | 46 | 37 | job, work, project |
| **Sports** | | | |
| Fitness | 45 | 34 | gym, exercise, fitness |
| Walking | 63 | 42 | walk, walking, going |
| Outdoor Activity | 37 | 22 | ride, climbing, cycling |
| Swimming | 50 | 29 | swimming, pool, sea |
| **Health** | | | |
| Hospital | 98 | 60 | hospital, operation, pain |
| Mental Therapy | 28 | 17 | mental, therapy, NHS |
| Sleep | 36 | 27 | sleep, tired, rest |
| **Other Themes** | | | |
| Pet | 73 | 49 | dog, puppy, cat |
| Weather | 86 | 54 | sunshine, rain, warm |
| House | 90 | 62 | house, decorating, bedroom |
| Covid-19 | 132 | 82 | Covid-19, virus, vaccine |
| Haircut | 23 | 21 | haircut, hair, cut |